\documentclass[letterpaper, journal]{IEEEtran} 

\IEEEoverridecommandlockouts

\usepackage{amsmath} 
\usepackage{circledsteps}
\usepackage{newtxtext,newtxmath}
\usepackage{xspace}
\usepackage{caption}
\usepackage{subcaption}
\usepackage{graphicx}
\usepackage{float}
\usepackage{cite}
\usepackage{placeins}
\usepackage{hyperref}
\usepackage{multirow}
\usepackage{booktabs}

\newcommand{\DDPGx}{$\text{DDPG}\land$}

\newcommand{\firmware}{SwaNNFlight}
\newcommand{\framework}{SwaNNL}
\newcommand{\frameworklong}{SwaNN Lake}
\newcommand{\rev}[1]{\textcolor{black}{#1}}
\newcommand{\n}[1]{#1)}

\DeclareMathOperator*{\argmax}{arg\,max}
\DeclareMathOperator*{\argmin}{arg\,min}

\newcommand{\pand}[1]{\ensuremath{\boldsymbol{\land}^{#1}}\xspace}

\title{\LARGE \bf
Sim-Anchored Learning for On-the-Fly Adaptation
}

\author{Bassel El Mabsout$^{1}$, Siddharth Mysore$^{1}$, Shahin Roozkhosh$^{1}$, Kate Saenko$^{1}$ and Renato Mancuso$^{1}$
\thanks{$^{1}$Affiliated with the Computer Science department at Boston University}%
}

\newcommand{\piSource}{\ensuremath{\pi_{\text{S}}}}

\newcommand{\piAnchor}{\ensuremath{\pi_{\text{S} \overset{\smash{\psi}}{\triangleright}\text{T}}}}
\newcommand{\piNaive}{\ensuremath{\pi_{\text{S}\triangleright\text{T}}}}

\begin{document}

\maketitle
\thispagestyle{empty}
\pagestyle{empty}


\section{Introduction}
    Reinforcement Learning (RL) offers powerful solutions for complex control problems where traditional techniques may be inadequate.
    Due to the expensive and potentially risky nature of real-world robot interaction, RL agents are typically trained and evaluated in simulated environments.
    However, recent work has shown that controllers applied to real systems can exhibit undesirable properties, from minor control issues to severe hardware failures~\cite{benchmarkingRobo, Sim2Real, Sim2multi}.
    These problems stem from the challenge of crossing the domain gap between simulated training environments and the real world---the ``reality gap''.

    While direct on-robot training raises safety and cost concerns, making simulated training attractive, the transition from simulation to reality presents significant challenges\cite{Muratore2022, pmlr-v155-sandha21a}.
    While the "reality gap" often focuses on dynamics discrepancies, a key related difficulty arises from the \textbf{distributional sim-to-real gap}.
    Real-world experience collection, especially during initial adaptation, frequently yields limited or skewed state distributions compared to the broader coverage achievable in simulation.
    Indeed, real-world control scenarios often generate distributions where critical scenarios occur only in the tails.
    Relying solely on such skewed real-world data means the learned RL value function (e.g., the \mbox{Q-value}) often fails to capture the designer's intended priorities.
    While reward functions are typically considered the primary specification of intent, we posit that comprehensive priorities are also implicitly encoded within the simulation through controllable choices like task setups and state initializations.
    This broader, simulation-encoded intent is precisely what risks being lost during adaptation on limited real-world data, leading to catastrophic forgetting.
    We propose using the value function derived from this intentionally designed simulation as an ``\textbf{anchor for intentionality}'' during real-world adaptation.
    Optimizing against both the anchor critic ($Q_\Psi$) and a real-world critic ($Q_\pi$) allows adaptation while preserving crucial, intentionally designed behaviors.
    This facilitates robust sim-to-real transfer that respects the intended operational goals across the full state space.

    Implementing and evaluating adaptation strategies like anchor critics directly on resource-constrained robots presents significant practical challenges.
    While many existing solutions offload computations to ground-station servers~\cite{Chinchali2021,orevi2023}, this can limit autonomy and real-time responsiveness.
    To facilitate the real-world validation of our anchor critic approach and enable practical on-the-fly adaptation, we developed and contribute \firmware{}: an open-source firmware stack.
    By enabling efficient on-board inference coupled with ground-station communication for policy updates, \firmware{} provides the necessary infrastructure for the live adaptation experiments discussed herein, making robust, real-time control adjustments viable for real-world robotic applications.
\section{Related Work}\label{sec:related_work}

    Bridging the sim-to-real gap in RL requires addressing both dynamics discrepancies and the critical distributional shifts between simulation and reality highlighted previously \cite{Muratore2022, pmlr-v155-sandha21a}.
    Existing transfer strategies approach this challenge in various ways, but often struggle to adapt effectively without losing the comprehensive behavioral intent encoded within the simulation's design.

    \textbf{Sim-to-Real Transfer Strategies:}
    Techniques like domain randomization aim to improve robustness by varying simulation parameters, though fully capturing real-world variability can be challenging \cite{Muratore2022}.
    Domain adaptation methods fine-tune policies on real data but often suffer from \textbf{catastrophic forgetting}, where agents overfit to the limited target distribution and lose crucial behaviors learned across the simulation's broader context \cite{catastrophic-forgetting-wolczyk}.
    Our work specifically targets this forgetting problem arising from distributional shifts during adaptation.

    \textbf{Stable Adaptation Techniques:}
    Ensuring stability during adaptation is crucial.
    Methods like Trust Region Policy Optimization (TRPO) and Proximal Policy Optimization (PPO) \cite{PPO} limit policy updates to prevent divergence.
    However, these methods do not explicitly focus on preserving the specific, intentionally designed behaviors learned from the diverse scenarios in simulation.
    Our Anchor Critics approach complements stability concerns by directly anchoring the policy to the simulation-encoded value function ($Q_\Psi$) while adapting based on real-world experience ($Q_\pi$).

    \textbf{On-Robot Adaptation and Infrastructure:}
    Applying these strategies on resource-constrained robots presents practical hurdles.
    Prior work like Neuroflight demonstrated embedding NN controllers in firmware but faced challenges with sim-to-real transfer and update flexibility \cite{NFori, NFv2}.
    Our \firmware{} firmware stack builds upon this, enabling efficient on-board inference coupled with ground-station communication for live policy updates \cite{swannlake-github}.
    This provides the necessary platform for deploying advanced adaptation methods like Anchor Critics in real-time.

    \textbf{Positioning Relative to Prior Work:}
    While existing work addresses dynamics gaps and stability, our method uniquely tackles catastrophic forgetting induced by distributional shifts by anchoring adaptation to the simulation's comprehensive design intent.
    Combined with the \firmware{} infrastructure, we offer a practical and theoretically grounded solution for robust live adaptation.

\section{Live Adaptation with Anchor Critics}\label{subsec:anchor_crit}

\begin{figure}[H]
    \centering
    \includegraphics[width=\columnwidth]{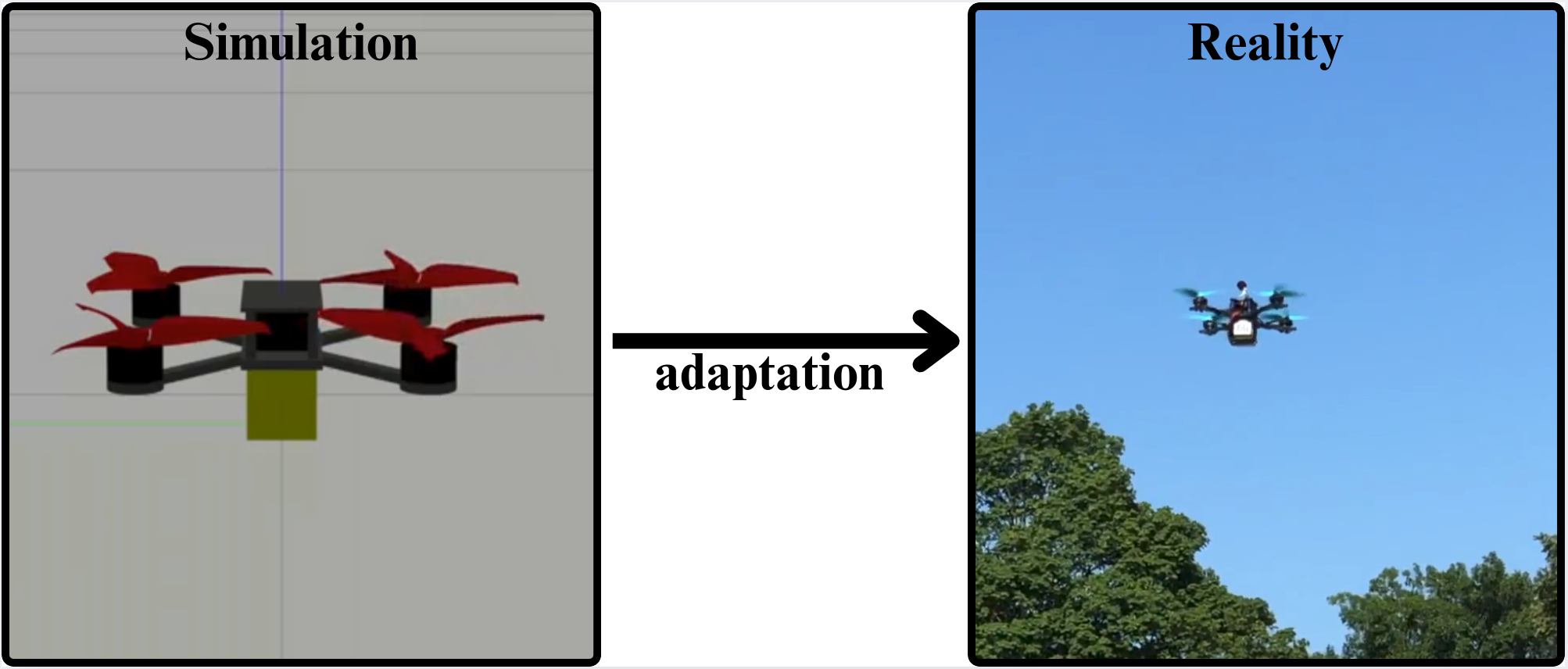}
    \label{fig:adaptation}
\end{figure}

    The main problem our work seeks to address is that of transferring learned control policies between similar domains in real-time.
    In our main use-case, this is transferring controllers trained in simulation to the real-world, where simulated dynamics may not be sufficiently accurately captured to enable sim-to-real transfer without performance degradation.
    A common issue of sim-to-real policy transfer is erratic control, which leads to high power consumption and motor over-stress.
    Our proposed solution attempts to bridge the domain gap during transfer by adapting controllers to the new domain during live operation, aiming to refine and enhance the control response.

    \paragraph*{Domain Gap}
        The `domain gap' involves dynamics changes, and attempting adaptation often leads to \emph{catastrophic forgetting} where simulation-learned behavior is lost, causing instability \cite{catastrophic-forgetting-wolczyk, catastrophic-forgetting-binici}.
        Furthermore, the distribution of states encountered during real-world operation can be highly skewed, potentially underrepresenting critical scenarios (like an emergency maneuver).
        Optimizing solely on this skewed distribution risks learning policies that forget the broader priorities encoded across the simulation's design (including tasks, initializations, and rewards).
        When new observations are not consistent with simulated observations, unpredictably aberrant behavior is a safety risk.
        Unlike in simulation, it is not possible in real-life to safely initialize the control system in arbitrary and diverse conditions.
        Experience in the new domain needs to start being gathered from a `safe' region and boundaries of safety can be expanded as the policy adapts to the new domain.
        However, this induces a bias in the adapted policy towards those initial `safe' experiences -- and while some degree of bias is beneficial for the policy to adapt itself, balancing the effects of the new experience against the beneficial experiences gained in simulation can be tricky.

    \paragraph*{A Case Against Mixing Experience Buffers During Adaptation}\label{subsec:mixedBuffers}
        It may seem reasonable, as a first-pass solution, to adapt policies by mixing simulated and real experience during fine-tuning.
        However, this can be problematic in two critical ways.
        (i) It breaks the Markov assumption via unobserved hidden variables about the data's domain origin, introducing inconsistency.
        (ii) Ensuring a meaningful balance between potentially skewed real-world data and intentionally curated simulation data is difficult, risking reward skew (Section~\ref{subsec:data_skew}).
        Instead, we suggest that separating the value functions that drive policy optimization between domains would enable weighing how much policies 'care' about one domain over the other.
        Thus allowing a tunable trade-off of stability over adaptability and vice versa.

    \paragraph*{Anchor-Critics}
        To mitigate the effects of domain transfer and adapting to a new domain live during active deployment, we introduce \emph{Anchor-Critics} (more details in Section~\ref{subsec:anchor_critics}).
        Anchors are an augmentation to typical Actor-Critic optimization, where we keep the value function trained on the initial (e.g. simulated) domain during policy fine-tuning to `anchor' the policy to its previously learned behavior, and train a secondary, independent critic to reflect value estimation on the target domain (e.g. real-world).
        The balance between the anchor and the new critic is tunable, and consequently, how much a policy is allowed to deviate from previously learned behavior during fine-tuning for adaptation is also tunable.
        We are able to take some of the guess-work out of experience balancing and ensure better stability by limiting policy drift.
        In the following sections, we will demonstrate the utility of anchors across a range of benchmark RL environments as well as for the motivating problem of live policy adaptation on a quadrotor drone.
        
    \subsection{Data Distribution Impact on Policy Optimization}\label{subsec:data_skew}
    
        Multi-objective optimization is significantly impacted by the coverage of the data upon which solutions are optimized.
        The value of simulation lies partly in the designer's ability to enforce broad data coverage, reflecting intended priorities through controllable choices like task setups and initializations, a luxury often unavailable in real-world data collection.
        Consider, for example, a two-objective policy optimization which aims to simultaneously minimize actuation (analogous to power consumption), $\phi_p$, and goal-tracking error, $\phi_e$, such that the optimal policy, $\pi^*$ is given by:
        \begin{equation}
            \pi^* = \argmin_{\pi} \left(f(\phi_p(s), \phi_e(s))\right) \ \ \forall s \in S
        \end{equation}
        for all states $s$ in the set of valid states and with objectives composed by some objective composition function $f$.
        With a sufficiently diverse set of training examples, $\hat{S} \approx S$, good coverage allows a policy optimized over $\hat{S}$ would be a reasonable approximation of the optimal policy.
        However, if coverage is limited to a limited region of the state-space such that one of the criteria becomes degenerate this could drive policies into trivial optima.
        For example, if a majority of the data involves a drone hovering in place, where $\phi_e \approx 0$), the optimal actuation would be no actuation, but this would also correspond to an abdication of control.
        This trivial solution is at odds with the requirements of controlability. A formalized example of this issue is given in Section~\ref{subsec:reward_skew}.
        
        In the specific example of high-performance racing drone control, where swift aerobatic response is critical, the safest control scenario in which a policy trained in simulation can be deployed is to first tune a policy for stable flight in slower maneuvers.
        This biases the fine-tuning data capture towards low actuation as new data is collected, even with past simulated experiences retained in a replay buffer as their proportion diminishes, and causes the policy to forget more agile control.
        As actor-critic methods make use of replay buffers, the issue of forgetting appears in various forms accross multiple tasks we explore on a real quadrotor in Section~\ref{subsec:forgettingAndUnsafe} and on benchmark control tasks in Section~\ref{subsec:simAnchors}.
        \subsection{Illustrating Reward Skew}\label{subsec:reward_skew}

        Consider a reward function $R(s, a)$ linearly composed of two components, both bounded between 0 and 1 such as
        \begin{equation}
            R(s, a) = w_p R_p(s, a) + w_d R_d(s, a),
        \end{equation}
        where $R_p$ is the reward component for power efficiency, $R_d$ is the reward component for distance to target, and $w_p, w_d$ are their respective weights.
        
        Let us separately define $R_p$ and $R_d$ as follows:
        \begin{equation}
            R_p(s, a) = 1 - \frac{\|a\|^2}{\|a_\text{max}\|^2},
        \end{equation}
        where $a$ is the action vector and $a_\text{max}$ is the maximum possible action vector. Smaller actions are considered more power efficient in this scenario. Next, we define
        \begin{equation}
            R_d(s, a) = 1 - \frac{\|s - s_\text{setpoint}\|}{\|s_\text{max} - s_\text{setpoint}\|},
        \end{equation}
        where $\|s - s_\text{setpoint}\|$ is the distance between the current state and the setpoint. $R_d(s,a)$ therefore represents how near the quadrotor is to the target position and orientation.
        
        By substituting our reward function and separating the expectations, we can define
        \begin{equation}
            Q^\pi(s, a) = w_p Q_p^\pi(s, a) + w_d Q_d^\pi(s, a),
        \end{equation}
        where $Q_p^\pi(s, a)$ and $Q_d^\pi(s, a)$ are the Q-values for power efficiency and distance to target, respectively.
        
        Now, consider two scenarios:

        1) The environment distribution is heavily skewed towards states near the target---as is the case when the quadrotor is hovering.
            In this scenario, $\|s - s_\text{setpoint}\| \approx 0$, which implies $R_d(s, a) \approx 1$. Consequently we have
            \begin{equation}
                Q_d^\pi(s, a) = \mathbb{E}_{\tau \sim \pi}\left[\sum_{t=0}^{\infty} \gamma^t \cdot 1 | s_0 = s, a_0 = a\right] \approx \frac{1}{1-\gamma}.
            \end{equation}
            This is the maximum possible value, as $R_d$ is always near 1 in this scenario. The final Q-value and optimal policy becomes
            \begin{equation}
                \begin{aligned}
                Q^\pi(s, a) &\approx w_p Q_p^\pi(s, a) + \frac{w_d}{1-\gamma}\\
                \pi^* &= \argmax_{\pi} w_p V_p^\pi(s).
                \end{aligned}
            \end{equation}
        
            This policy will focus solely on minimizing power consumption, as the distance to target term is constant and already maximized. Thus, the optimal policy will produce $||a||^2 = 0$, which is to perform no control at all.
        
        2) The environment has a more balanced distribution of states. As such, the optimal policy $\pi^*$ will remain
            \begin{equation}
                \pi^* = \argmax_{\pi} \left(w_p V_p^\pi(s) + w_d V_d^\pi(s)\right).
            \end{equation}
            This optimal policy will balance both power efficiency and trajectory tracking, depending on the relative weights $w_p$ and $w_d$.  
        This example highlights how reliance on skewed data can misdirect policy optimization. Our proposed anchor critic mechanism, detailed next, directly addresses this by incorporating the value function derived from an intentionally balanced simulation design, thereby preserving sensitivity to all intended objectives.

    \subsection{Anchor Critics}\label{subsec:anchor_critics}

    Our results suggest that the issue of catastrophic forgetting naturally arises in a wide range of robot controllers when live adaptation with RL is na\"ively performed.
    We observed catastrophic forgetting within only a few updates across various benchmark control problems, underscoring the severity of the issue.
    Simulation provides a controlled environment where diverse maneuvers can be safely trained, allowing for the design of tasks, rewards, and state distributions that collectively encode the desired operational intent. 
    Conversely, real-world data collection may be limited by safety constraints and yield skewed distributions unrepresentative of the full intended operating range.
    Valid live adaptation must therefore incorporate real data while respecting the comprehensive behavioral profile captured by the simulation's design.
    We term the domain in which the agent is initially trained (typically simulation) the \textbf{source domain} and the domain in which the agent is adapted (typically reality) the \textbf{target domain}.
    To tackle the challenges presented, we propose adapting agents based on two main value criteria: (i) the \textbf{Q-value $Q_{\pi}$ learned on data observed from the target domain}, reflecting adaptation to current dynamics and the observed data distribution, and (ii) the \textbf{Q-value $Q_{\Psi}$ learned on data from the source domain}, representing the performance according to the simulation-encoded intent.
    We term the critic providing $Q_{\Psi}$ the \textbf{`anchor' critic}, as it anchors the policy optimization to the intended behavior profile shaped by the deliberate simulation design choices (tasks, initializations, rewards), preventing deviations caused by potentially limited or skewed target domain data.

    We formulate the adaptation challenge as a multi-objective problem: the agent must simultaneously satisfy performance objectives derived from the source domain's design intent and objectives learned from the target domain's experiences. Following Fulfillment Priority Logic (FPL)~\cite{fpl2025}, we represent these objectives as \emph{fulfillments} in the range $[0,1]$.
    Assuming rewards are normalized such that Q-values can be appropriately scaled to represent expected fulfillment within $[0,1]$ (e.g., $Q(s,a) \times (1-\gamma)$), we can apply FPL composition operators. For notational simplicity, we will use $Q$ to denote these fulfillment-compatible values representing the expected performance in the source ($Q_{\Psi}$) and target ($Q_{\pi_\theta}$) domains.
    Our goal is to achieve high simultaneous fulfillment across both domains. To compose these Q-values for joint satisfaction, we utilize the FPL conjunction operator $\pand{0}$ (geometric mean), as defined in FPL~\cite{fpl2025}, applied to the target fulfillment $Q_{\pi_\theta}$ and the anchor fulfillment raised to the priority weight $w_\Psi$, i.e., $Q_{\Psi}^{w_\Psi}$.
    The geometric mean, a specific case ($p=0$) of FPL's power mean conjunction, inherently rewards joint satisfaction---the value is high only if both $Q_{\pi_\theta}$ and the prioritized $Q_{\Psi}$ are high---and implicitly prioritizes the improvement of less-fulfilled objectives, consistent with FPL's goal of balancing competing goals.

    The anchored policy optimization goal, given anchor critic $Q_{\Psi}$, thus modifies the standard actor loss $J_{\pi_\theta}$ to maximize this geometric mean composition $J_{\pi_\theta}^{\Psi}$:
    {
        \begin{equation}
             \mathbb{E}_{s_r \sim \mathcal{D}_{\text{T}}, s_s \sim \mathcal{D}_{\text{S}}}[ Q_{\pi_\theta}(s_r, \pi_\theta(s_r)) \pand{0} (Q_{\Psi}(s_s, \pi_\theta(s_s))^{w_\Psi}) ]
        \end{equation}
    }
    where $\mathcal{D}_{\text{T}}$ and $\mathcal{D}_{\text{S}}$ are the replay buffers for the target and source domains respectively. The exponent $w_\Psi$ acts as a user-defined \textbf{priority weight}. While FPL uses a specific syntax ($[\phi]_\delta$) for explicit prioritization, $w_\Psi$ serves the analogous purpose here, allowing the user to control the relative importance of fulfilling the source domain's design intent ($Q_{\Psi}$) versus the target domain's observed objectives ($Q_{\pi_\theta}$). A higher $w_\Psi$ prioritizes preserving the original simulation behavior.
    
    To keep anchors representative on the source domain, we keep updating $Q_{\Psi}$ (and thus $Q_{\Psi}$) during training, following the typical actor-critic update but using only data sampled from $\mathcal{D}_{\text{S}}$.
    The Q-value $Q_{\pi_\theta}$ (and $Q_{\pi_\theta}$) is conversely trained only on data sampled from $\mathcal{D}_{\text{T}}$.
    
    This simultaneous sampling and optimization approach ensures that the policy maintains a balance between adapting to the target domain data while preserving the knowledge gained from source domain training, effectively mitigating catastrophic forgetting.

    \begin{figure*}[t]
        \centering
        \begin{subfigure}[h]{0.32\textwidth}
            \centering
            \includegraphics[width=\textwidth]{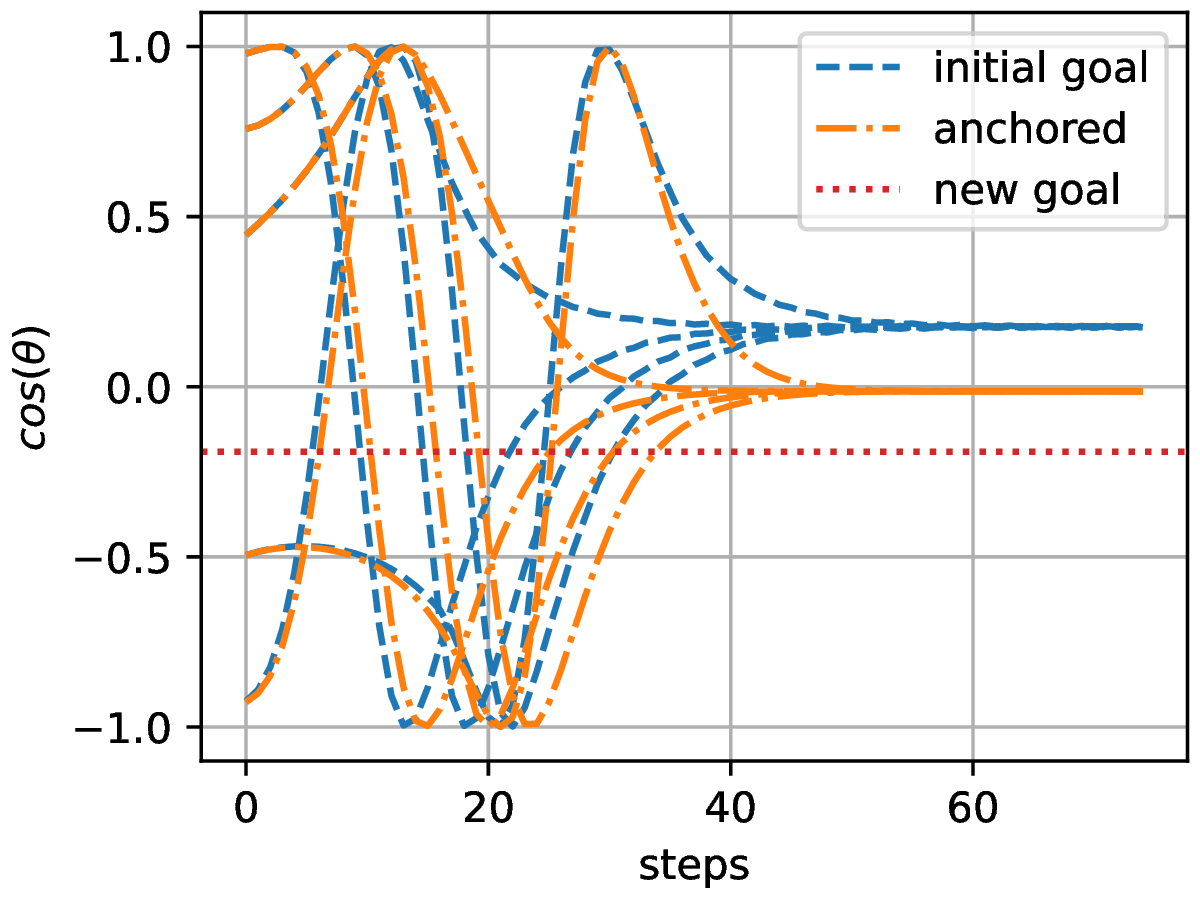}
            \caption{DDPG}
            \label{fig:pendulum_ddpg}
        \end{subfigure}
        \begin{subfigure}[h]{0.32\textwidth}
            \centering
            \includegraphics[width=\textwidth]{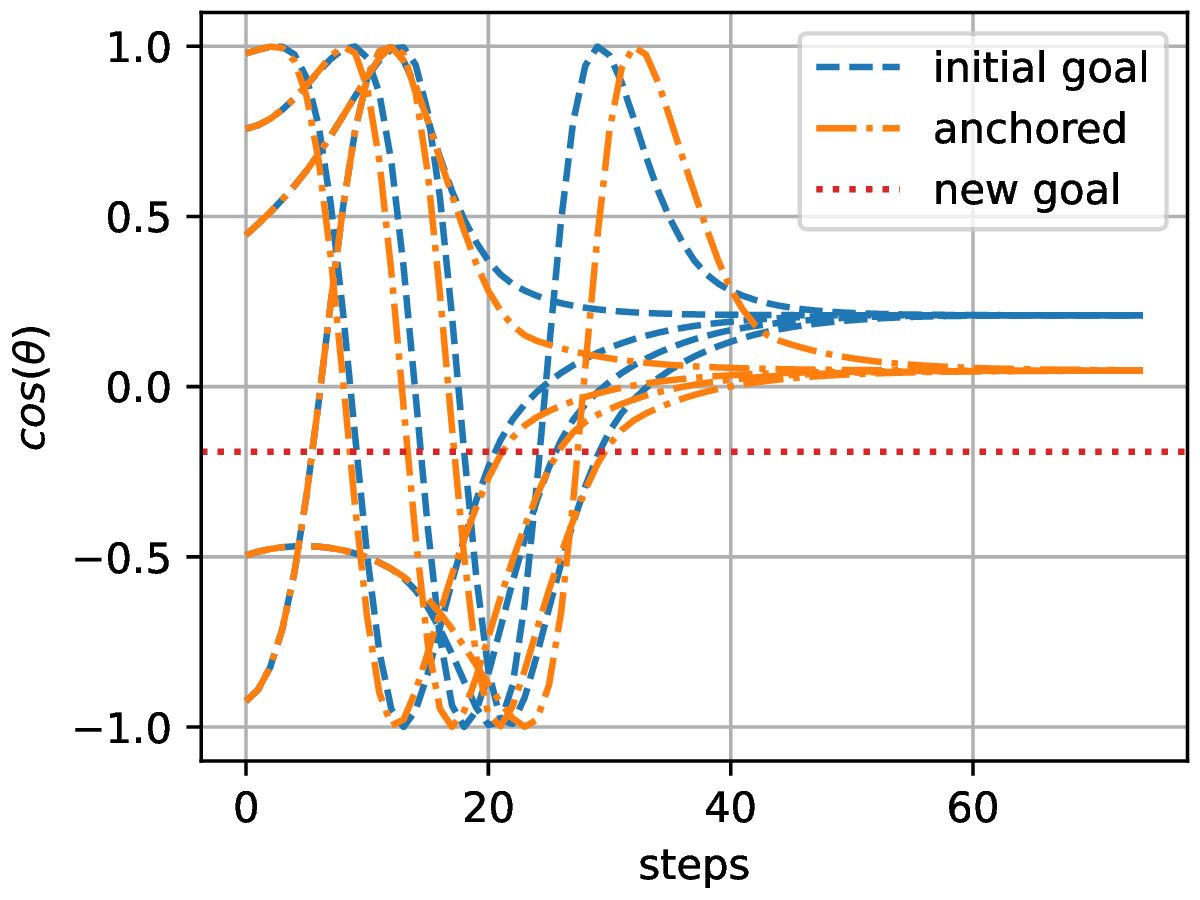}
            \caption{SAC}
            \label{fig:pendulum_sac}
        \end{subfigure}
        \begin{subfigure}[h]{0.32\textwidth}
            \centering
            \includegraphics[width=\textwidth]{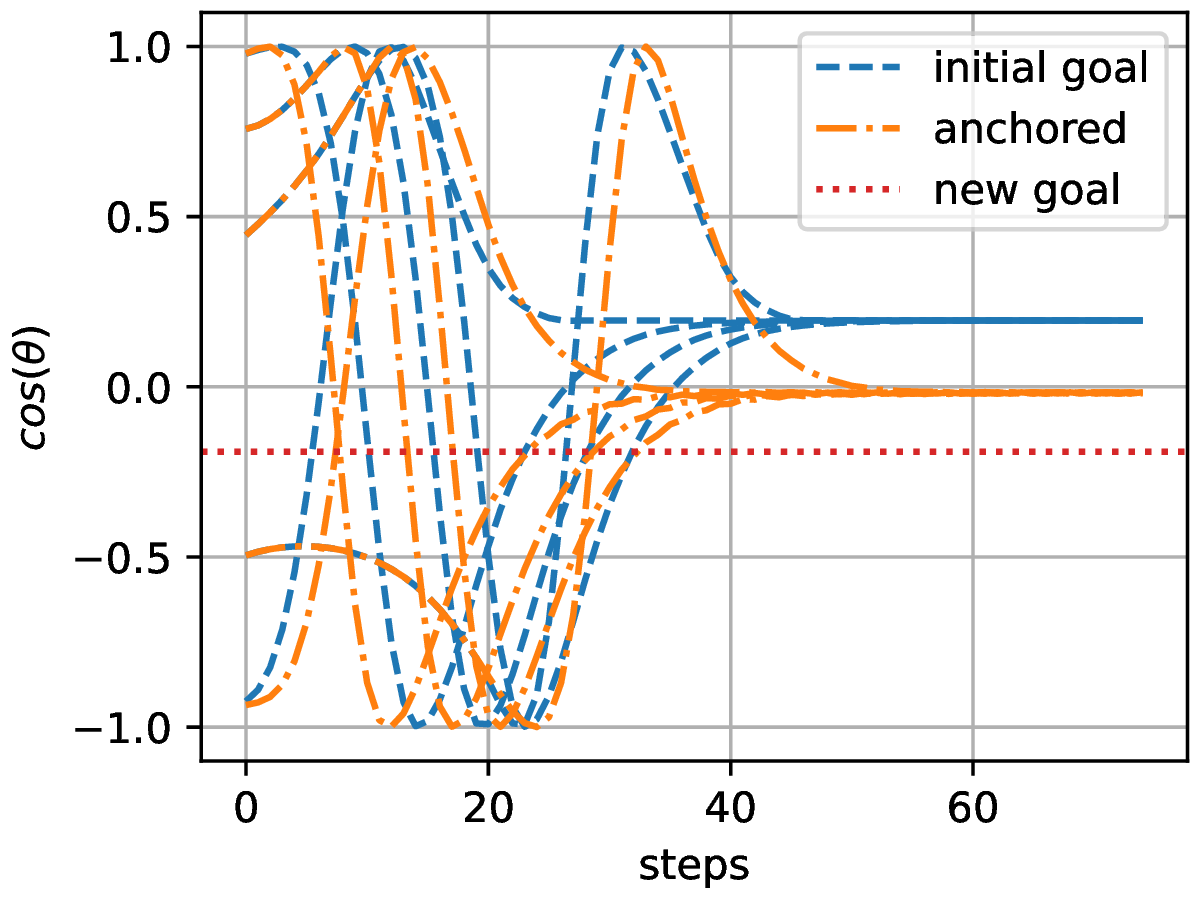}
            \caption{TD3}
            \label{fig:pendulum_td3}
        \end{subfigure}
        \caption{\rev{Evolution of the angle ($sin(\theta)$) of an inverted pendulum over time for policies trained with anchor-critics tested on DDPG, SAC and TD3.
        There are 5 different test runs per goal, and the angle of the pendulum is initialized at random.
        All 3 anchored algorithms consistently learn to point to 0 rad, thus maintaining a good compromise between our initial goal and our new goal.}}
        \label{fig:pendulum_all}
    \end{figure*}    
\section{\emph{Anchor Critics in Simulation}}\label{sec:simAnchors}

    In this section, we present a comprehensive evaluation of anchor critics through a series of experiments. First, we demonstrate the general applicability of anchor critics on a toy problem, detailed in Section~\ref{subsec:invPend}. Our initial explorations provide insights into the basic functionality and potential of anchor critics.

    Following this, we conduct sim-to-sim transfer evaluations using modified Gymnasium environments in Section~\ref{subsec:simAnchors}. These modifications, which introduce shifts in dynamics or task distributions, reflect specific adaptation goals. They allow us to showcase how such distributional shifts can induce catastrophic forgetting even without a reality gap, and demonstrate how anchor critics effectively mitigate this issue by preserving intended behaviors learned in the source domain.
    We provide anchors as a library and allow the experiments to be reproduced~\cite{swannlake-github}.

    \subsection{Testing Anchors on Inverted Pendulum}\label{subsec:invPend}

    In this and the following sections, we evaluate the behavior of agents refined using anchor critics when fine-tuning controllers to establish the effectiveness of the approach in terms of (i) impact on policy stability, (ii) catastrophic forgetting.
    
        \begin{figure}[H]
        \begin{subfigure}{.32\columnwidth}
          \centering
          \includegraphics[width=.35\textwidth]{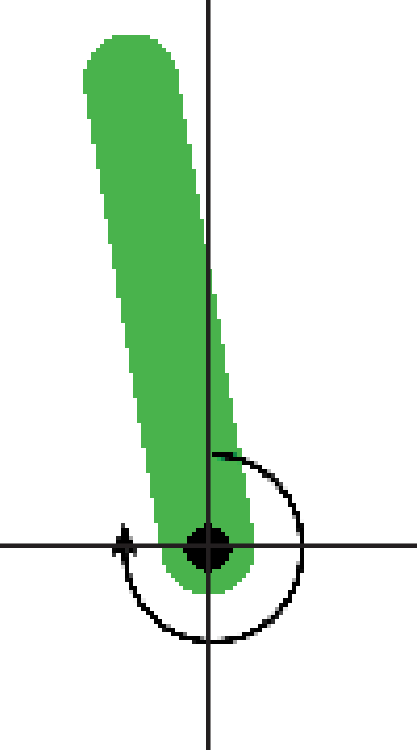}
          \caption{$\pi_\text{\tiny S}$ on source: \\ \hspace*{0.35cm} $\alpha = 10^\circ$}
          \label{fig:left_pendulum}
        \end{subfigure}%
        \begin{subfigure}{.32\columnwidth}
            \centering
            \includegraphics[width=.35\textwidth]{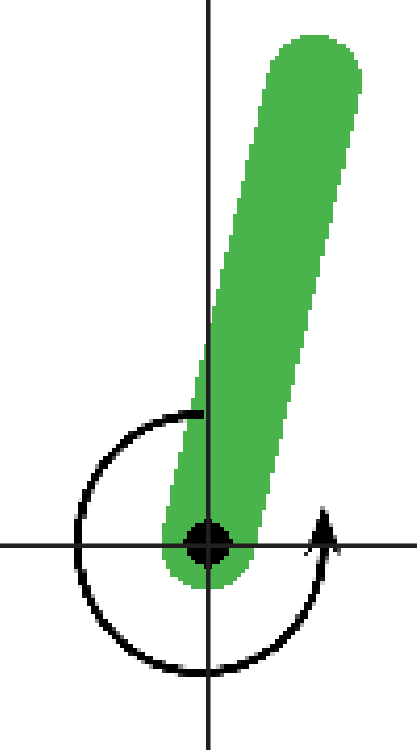}
            \caption{\piNaive{} on target: \\ \hspace*{0.35cm} $\alpha = -10^\circ$}
            \label{fig:right_pendulum}
        \end{subfigure}
        \begin{subfigure}{.32\columnwidth}
            \centering
            \includegraphics[width=.35\textwidth]{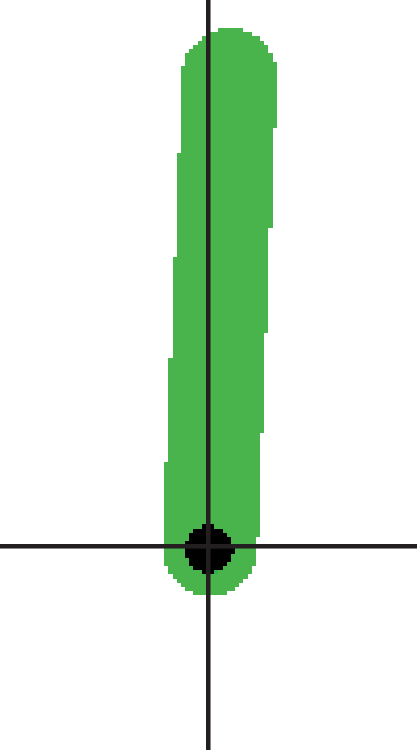}
            \caption{\piAnchor{} on target\\ \hspace*{0.35cm} with source anchor}
            \label{fig:up_pendulum}
        \end{subfigure}
        \caption{
            Policy \piSource{} in (a) is trained in the source domain where the goal is to hold the pendulum stably leaning to the left. We then fine tune \piSource{} on the target domain where the goal is to lean the pendulum to the right, producing \piNaive{} in (b). We then fine-tune \piSource{} again on the target domain but this time anchored on the source domain. This policy, termed \piAnchor{}, is shown in (c). \piAnchor{} holds the pendulum straight-up with no significant lean indicating that anchors find a compromise between the source and target domains.
        }
        \label{fig:pendulum_overview}
        \end{figure}

        \begin{figure}[H]
            \centering
            \includegraphics[width=\columnwidth]{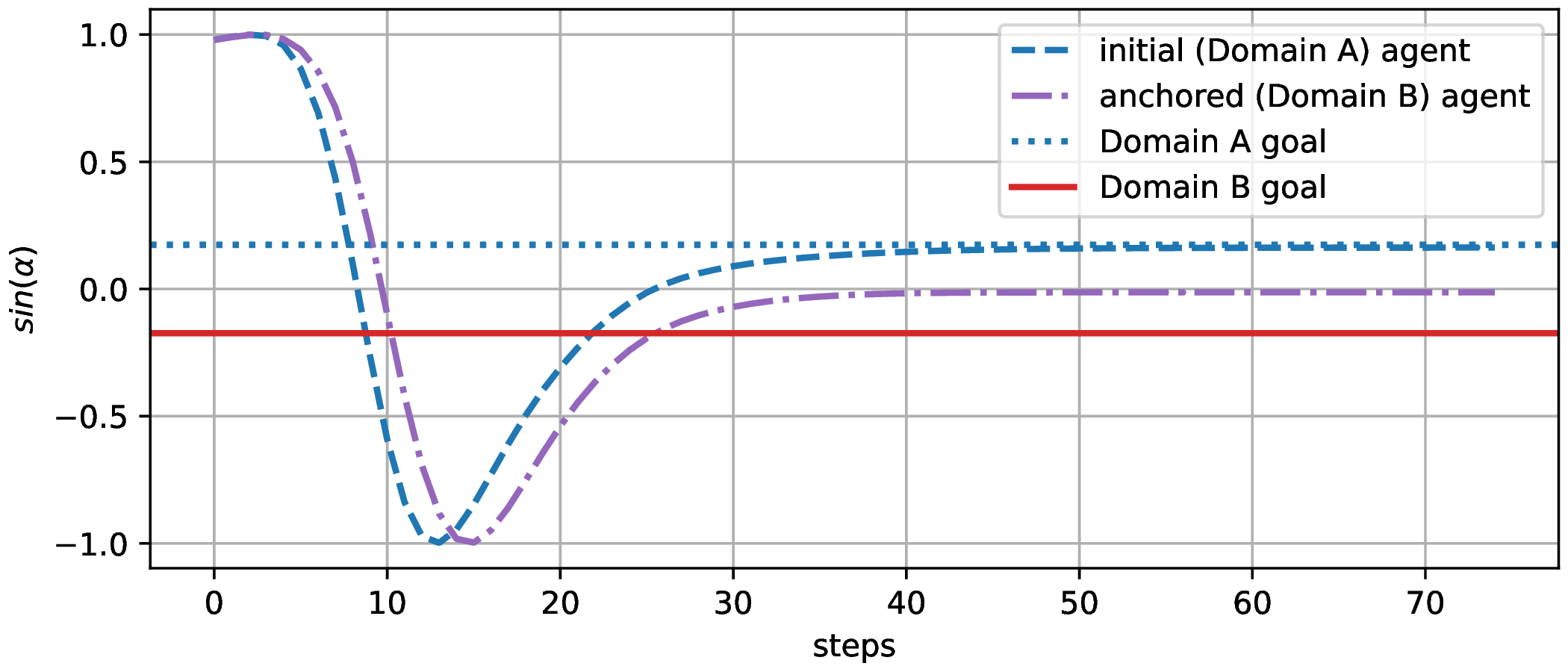}
            \caption{
                The sin of the inverted pendulum's angle $\alpha$ over time starting from $\alpha = 90^\circ$ for a DDPG agent trained initially on Domain~A as per Fig.~\ref{fig:left_pendulum}, and then adapted to Domain~B with Domain~A anchors as per Fig.~\ref{fig:up_pendulum}.
                Note that initial training allows the agent to stably balance the pendulum to approximately $10^\circ$ while adaptation results in the agents driving the pendulum to approximately $0^\circ$.
            }
            \label{fig:ddpg_anchor_pendulum}
            \vspace{-\baselineskip}
        \end{figure}

        To test the general viability of anchor critics, we implemented them in three popular contemporary RL algorithms:~DDPG~\cite{DDPG}, SAC~\cite{SAC}, and TD3~\cite{TD3} in Fig.~\ref{fig:pendulum_all} and tested them on a modified inverted pendulum problem.
        The typical goal of the task is to stabilize a torque-controlled pendulum vertically upright, but we instead aim to hold the pendulum at a specified angle.
        As shown in Fig.~\ref{fig:pendulum_overview}, the naively fine-tuned agent exclusively solves the target domain task, forgetting the source behavior.
        The anchor critic, however, represents the value function associated with the source domain's intended behavior (leaning left). By incorporating this anchor, the agent is encouraged to find a policy that balances the conflicting target objective (leaning right) with the original intent, resulting in a compromise behavior (leaning straight up) as shown in Fig.~\ref{fig:up_pendulum}.
        The evolution of the pendulum angle under anchored adaptation is shown in Fig.~\ref{fig:ddpg_anchor_pendulum}.
        Additional data for more agents and for agents trained with different algorithms are available in Section~\ref{A:hypersearch}.

    \begin{figure}[H]
        \centering
        \includegraphics[width=\linewidth]{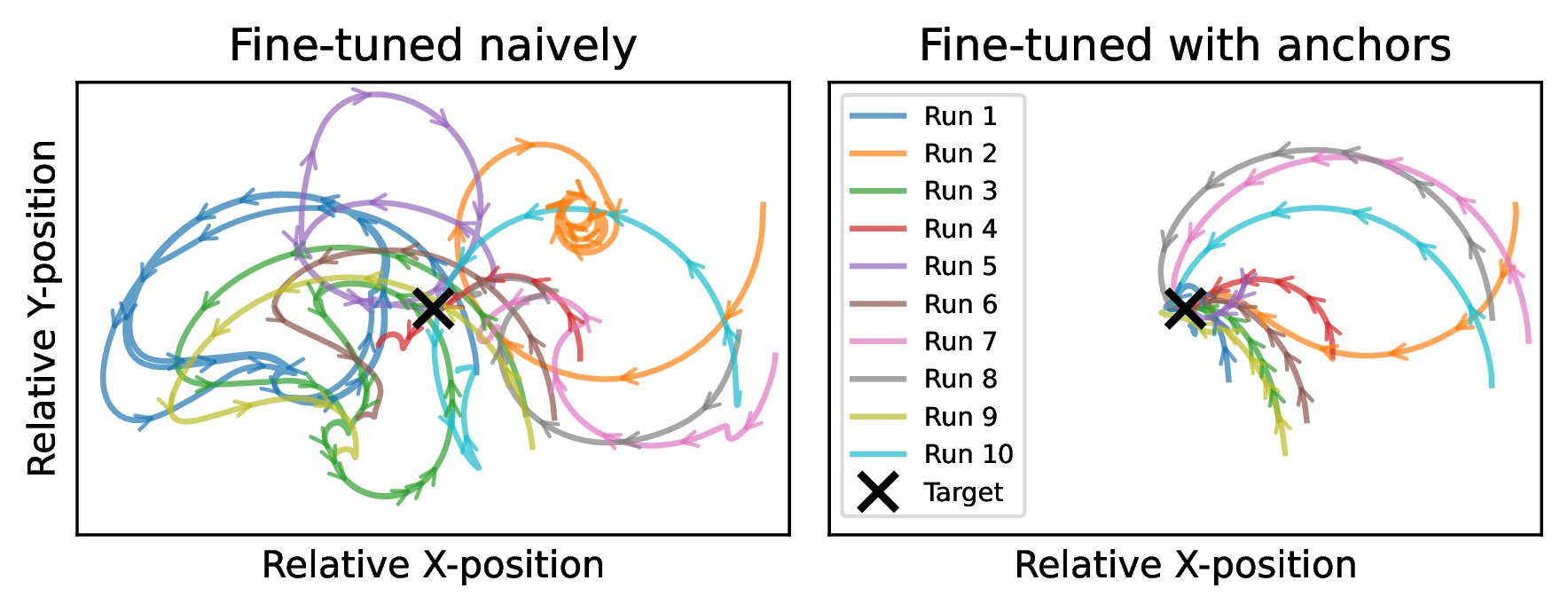}
        \caption{Evolution of Reacher agents on the source domain over time (arrows showing directionality). We contrast a naively fine-tuned agent with one that has been anchored. The plots show how unstable the naively tuned agent is, while the anchored agent is able to maintain stability.}
        \label{fig:reacher_evolution}
    \end{figure}
    \subsection{Catastrophic Forgetting in Reacher}\label{subsec:catastrophic_forgetting_reacher}
    This issue is especially grave on the reacher task. Here, the target domain has identical dynamics to the source, differing only in its goal distribution: target setpoints are restricted to a strict subset of the source domain's broader range (Table~\ref{tab:domain_specs}). This creates a limited, skewed task distribution during fine-tuning.
    The resulting instability of the naively fine-tuned agent (Fig.~\ref{fig:reacher_evolution}) stems from overfitting to this limited target distribution and consequently forgetting how to handle the wider range of goals present in the source domain.

    \begin{table}[H]
        \centering
        \begin{tabular}{|l|l|l|}
        \hline
        Environment & Source Domain & Target Domain \\
        \hline
        Pendulum-v0 & gravity 13m/s$^2$ & gravity 4m/s$^2$ \\
        Reacher-v4 & goal within 0.2m & goal within 0.1m \\
        LunarLanderContinuous-v2 & gravity 10m/s$^2$ & gravity 2m/s$^2$ \\
        \hline
        \end{tabular}
        \caption{Source and target domains for each environment.}
        \label{tab:domain_specs}
    \end{table}

    \begin{figure}[H]
        \centering
        \includegraphics[width=\linewidth]{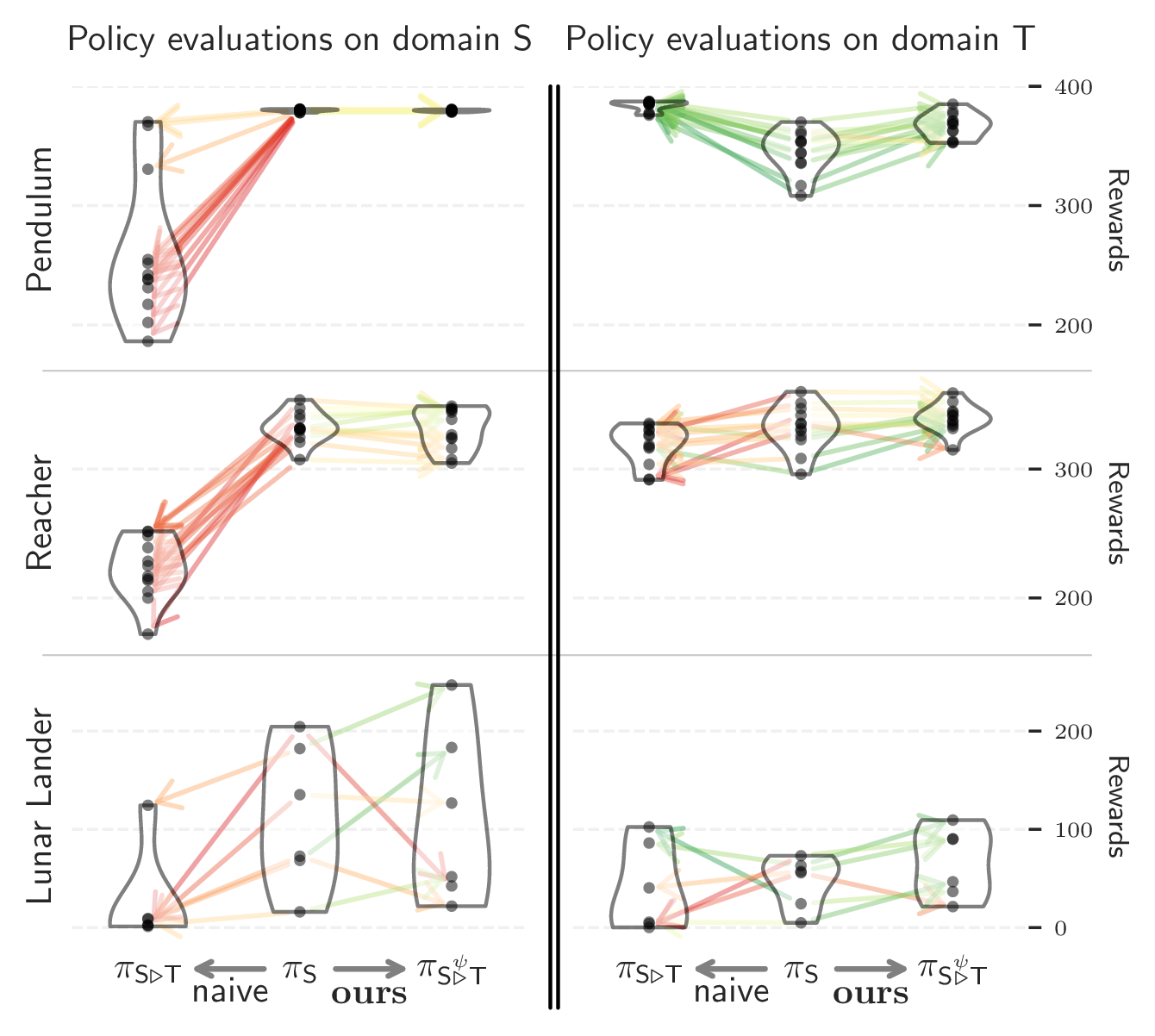}
        \caption{
            Distribution of rewards for agents evaluated on source (S) and target (T) domains. Each black dot indicates an agent's performance under a different random seed.
            Colored arrows indicate each agent's performance before (\piSource{}) and after fine-tuning na\"ively (\piNaive{}) or with anchors (\piAnchor{}), with green indicating an increase in performance and red indicating a decrease.
        }
        \label{fig:combined_results}
    \end{figure}
    \subsection{Preventing Catastrophic Forgetting on Gymnasium}\label{subsec:simAnchors}
        We perform experiments on three benchmark Gymnasium environments, namely Reacher-v4, LunarLanderContinuous-v2, and Pendulum-v0.
        For each environment, we vary some parameters as described in Table~\ref{tab:domain_specs} to test the impact of fine-tuning when a domain shift occurs.

    We then show that anchors can mitigate this issue accross multiple environments with reality-gaps both in dynamics and observation distributions in Fig.~\ref{fig:combined_results}.

    Our findings suggest that catastrophic forgetting occurs frequently when fine-tuning agents across domain shifts, where changes in dynamics or data distribution cause agents to forget previously learned behaviors.
    We find that anchors mitigate these issues. As seen in Fig.~\ref{fig:combined_results}, naively fine-tuned agents (\piNaive{}) degrade significantly on the source domain, likely overfitting to the target domain's specific distribution or dynamics and forgetting the broader intended behaviors.
    Anchored agents (\piAnchor{}), conversely, successfully adapt to the target domain while retaining source domain performance. This demonstrates their ability to maintain the balance intended by the source simulation design, even when faced with domain shifts.




\section{Anchor Critics in Reality}\label{sec:Eval}
    
    To address the challenges of consistent and safe live-adaptation with reinforcement learning in reality, we evaluate the effectiveness of anchor critics through a series of experiments:

    \begin{enumerate}
        \item Leveraging \framework{}, we evaluate how live-adapting with and without anchors affects safety, smoothness, and performance when adapting policies from simulation to a real drone system (Sections~\ref{subsec:forgettingAndUnsafe}~and~\ref{subsec:unconstrainedFlight}). We focus on how anchors mitigate issues arising from adapting to potentially limited real-world data distributions. The novel infrastructure offered by \framework{} enabled us to develop and test practical ways to adapt agents while a quadrotor is in flight, demonstrating how anchors make live adaptation safer and more robust against forgetting intended behaviors.

        \item We offer theoretical arguments motivating our method in Section~\ref{subsec:mixedBuffers}.

        \item Finally, we discuss known limitations with respect to drone control and the general cost of using anchors in Section~\ref{subsec:limitations}.
    \end{enumerate}

    \rev{Section~\ref{A:hypersearch} provides details on hyperparameters used, along with code for \firmware{} and \DDPGx{} RL training.}

    \subsection{Engineering \frameworklong{} for Live NN updates}\label{sec:firmware}
        \begin{figure}[H]
            \centering
            \includegraphics[width=\columnwidth]{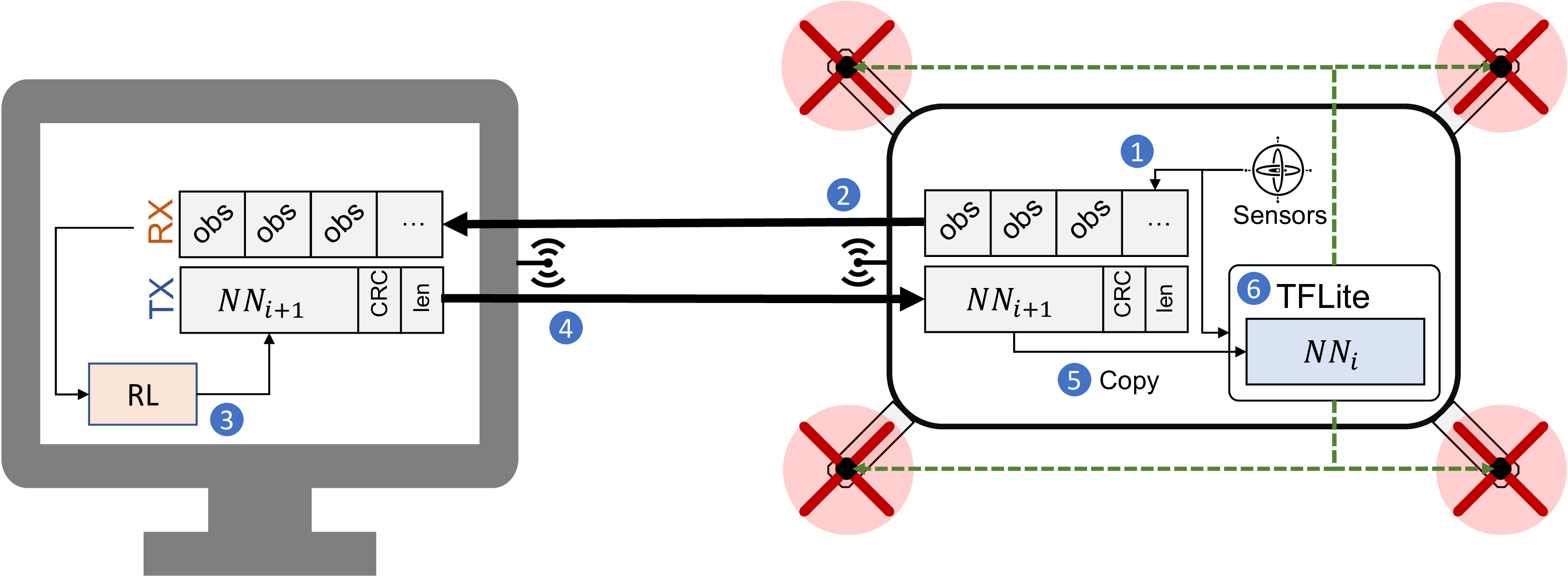}
            \caption{\framework{} overview for live adaptation. \n{1} Observations received as a result of policy invocation are packaged and \n{2} sent to the ground-station where they are unpacked and fed into an RL algorithm. \n{3} The updated policy is packaged with integrity check markers and \n{4} transmitted to the robot. \n{5} Upon verification of the received policy graph, the new graph is copied to system memory and \n{6} interpreted by TFLite for invocation. }
            \label{fig:sys_overview}
        \end{figure}
        In this work, we investigate strategies for RL-based live control adaptation on real robots. As such, two important sub-goals are formulated as follows. First, we needed to allow BetaFlight-compatible firmware to update its NN controller(s) without interrupting the control loop. As BetaFlight is a standard across various robotic platforms, this allows our framework to support a large range of robots, such as blimps, boats, helicopters, racing drones, and more. Second, we needed to be able adapt the agents live without crashing. This necessitated preventing catastrophic forgetting, particularly given the risk of overfitting to potentially skewed or limited data distributions encountered during live operation.
        
        We thus decided to evolve Neuroflight~\cite{NFori, NFv2}, initially developed by Koch~et~al. for allowing the use of NNs for high-performance quadrotor control.
        Neuroflight extends the open-source Betaflight flight-control firmware stack allowing NN models to be compiled and embedded into the control firmware.
        We evolved the Neuroflight framework such that modifications to the employed NN controller(s) can be carried out without interrupting the control loop.
        To do so, we re-designed the firmware so that we could swap out binarized NN graphs with minimal overhead rather than having them directly compiled into the firmware.
        In particular, it is possible to both (i) update the NN controller weights and (ii)~swap-in graphs with different NN architectures.
        Our novel firmware stack outlined in Fig.~\ref{fig:sys_overview}, is open-sourced~\cite{swannlake-github}.
        Our architecture allows inference to be performed onboard the micro-controller, but training onboard is prohibitively expensive, given compute and storage limitations of low-power embedded platforms.
        We thus introduce a point-to-point communication channel with an external ground station, to mitigate this issue.
        Most importantly, the ground station can be used as a training platform to adapt controllers based on data captured in flight.
        \firmware{} further integrates with wireless transceiver modules to (i) wirelessly transmit flight data and (ii) accept new NN models to switch to mid-flight.

        \section{TRAINING FLIGHT CONTROLLERS WITH \framework{}}

        Prior work employing the Neuroflight training framework trained RL-based controllers using TF1~\cite{NFori, NFv2, mysore2021train, mysore2021caps}.
        Attempting to deploy these RL-based controllers using our updated stack with TFLite's interpreter resulted in the control firmware crashing during invocation,
        \rev{due to the operations used by the translated TFv1 NNs.}
        Agents built with TF2, conversely, could be successfully deployed and invoked.
        This compelled us to update our training stack to TF2.
        
        Transitioning to TF2 introduced a problem for the training pipeline.
        Prior works \cite{mysore2021caps, mysore2021train, NFv2} train flight-control agents with the OpenAI Baselines implementation of PPO~\cite{PPO}.
        While reproducible with the software packages and versions reported by the authors, the performance observed with PPO agents in TF1 was not achievable by any of several tested TF2 implementations of PPO, including TF-agents, stable-baselines etc., despite using identical network architectures and performing a full hyperparameter search.
        Algorithms like PPO and TRPO are known to be sensitive to (and even brittle under) implementation specifics~\cite{henderson2018deep}, prompting a search for alternatives.
        
        Viable flight-control policies were trainable using a TF2 translation of OpenAI's Spinning Up PyTorch implementation of DDPG~\cite{DDPG}.
        While vanilla DDPG presented with similar oscillatory behavior identified in the work on CAPS ~\cite{mysore2021caps}, introducing CAPS regularization to DDPG enabled the training of viable agents in simulation.
        We found, however, that the performance of DDPG on the control problem at hand exhibited relatively high variance: not all the trained agents were safely flyable. This prompted us to explore alternative formulations which improve upon CAPS defined in the next section.

    \subsection{\rev{\firmware{} System Transceiving Metrics}}\label{A:comms_metrics}
        Wireless communication is handled by Digi XBee\textsuperscript{\circledR} ZigBee\textsuperscript{\circledR}-PRO radio-frequency modules. The drone communicates observation data to an XBee through a UART port, while the ground station uses an XBee to send updated network graphs back to the drone. This addition represents the only hardware change, with negligible impact on weight and power consumption.
    
        To ensure data integrity, we implemented a three-phase handshake protocol with cyclic redundancy checks (CRC). Upon verification, the drone atomically swaps to the new graph at the next control cycle. Buffered data is automatically chunked into CRC-validated packets, supporting multiple transmission rates.
    
        Using the MATEK-F722 controller with ZigBee\textsuperscript{\circledR}-PRO modules, we measured transmission and switching times for neural networks (NNs). Our control networks use two hidden layers with 32 neurons each. At a baudrate of 115200, sending an 8mb NN takes $\approx{11}$ seconds, with receiving handled in parallel to flight control. Switching to a new NN controller takes $\approx{134}$ ms.
    
        During flight, the drone transmits 59-byte state observations at a rate of 244 per second to the ground station.
        
    \subsection{Catastrophic Forgetting and Safety Concerns in Adaptation without Anchors}\label{subsec:forgettingAndUnsafe}
        
\begin{figure}[t]
            \centering
            \includegraphics[width=\columnwidth]{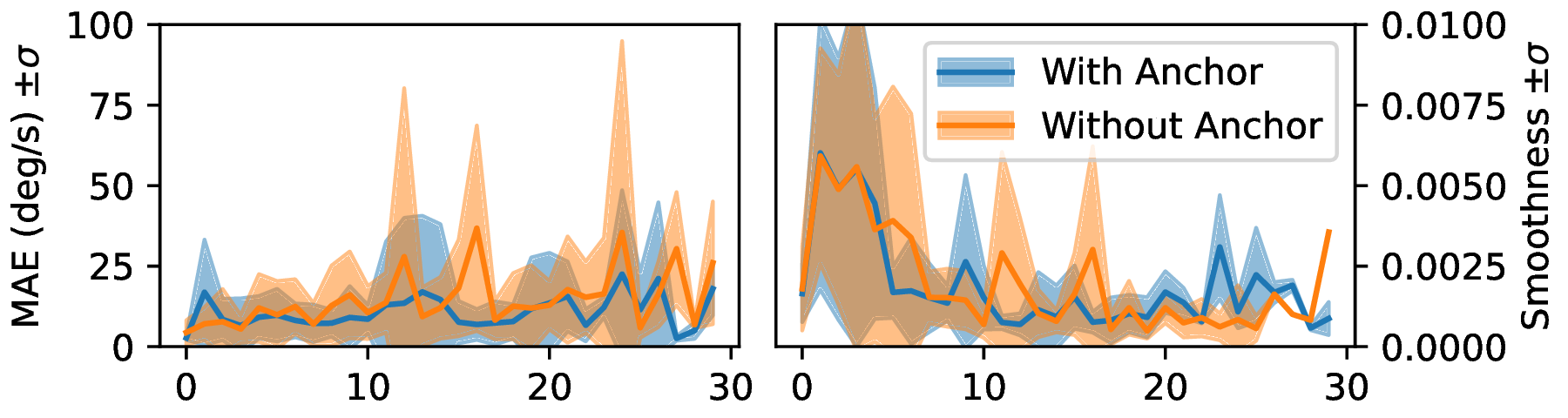}
            \caption{
                Shown here is how MAEs and smoothness evolve with adaptation steps, with and without anchors. 
                In both cases smoothness improves, however, without anchors, tracking errors spike often as catastrophic forgetting occurs. With anchors, error is bounded and remains in controllable range.
            }
            \label{fig:ac_vs_no_ac}
            \vspace{-.5\baselineskip}
        \end{figure}

\begin{figure}[t]
        \centering
        \includegraphics[width=\columnwidth]{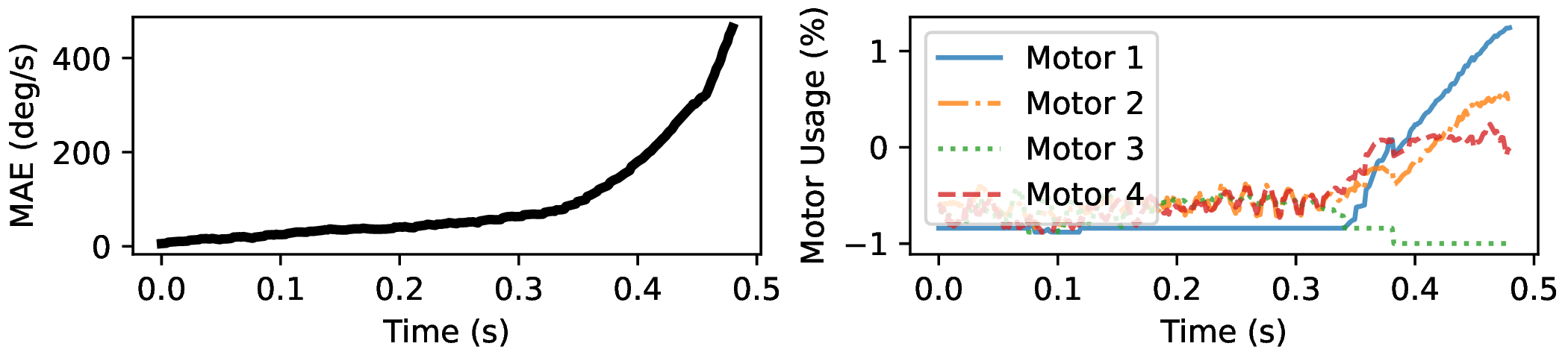}
        \caption{
            A snapshot of instability of the drone during an adaptation run without anchors.
            Observe how error increases exponentially as the drone controller becomes unstable.
        }
        \label{fig:instability}
        \vspace{-1.3\baselineskip}
    \end{figure}

        Before testing adaptation in live, unconstrained flight, we first conducted a series of tests in a controlled lab environment where the drone was attached to a light tether to restrain it in the event of a catastrophic loss of control, without being a hindrance to flight.
        These tests consisted of fine-tuning a well-trained controller from simulation (the source domain) on the real platform (the target domain) while giving it typically small control targets ($<50$ deg/s) and occasionally requesting control in excess of that ($>100$ deg/s).
        Without anchors preserving the agent's performance profile across the full range of control inputs defined in simulation, the live agents were observed to forget how to maintain tracking when receiving large inputs, as this occurs less frequently during adaptation. Such large inputs are common during acrobatic maneuvers, which are common in simulation tasks but not in real-world flights. We would observe agents sometimes regaining control, and often losing it again, as seen in Fig.~\ref{fig:ac_vs_no_ac}.
        More importantly, this also results in a sudden loss of control where errors would quickly snowball and motor actuation rises sharply, risking damage to the drone.
        Fig.~\ref{fig:instability} demonstrates how, when commanding attitudes not seen for a while during live-flight adaptation, agents could very quickly lose control (observed from MAEs increasing quickly and exponentially). We attributed this to overfitting to the limited or skewed distribution of the live-flight training data (predominantly small control inputs), similar to the Reacher environment. Anchors, by preserving the intended response learned across the diverse simulation distribution, help in avoiding this class of problems.
        
\begin{figure*}
    \centering
    \begin{subfigure}{.45\textwidth}
    \centering
    \includegraphics[width=\linewidth]{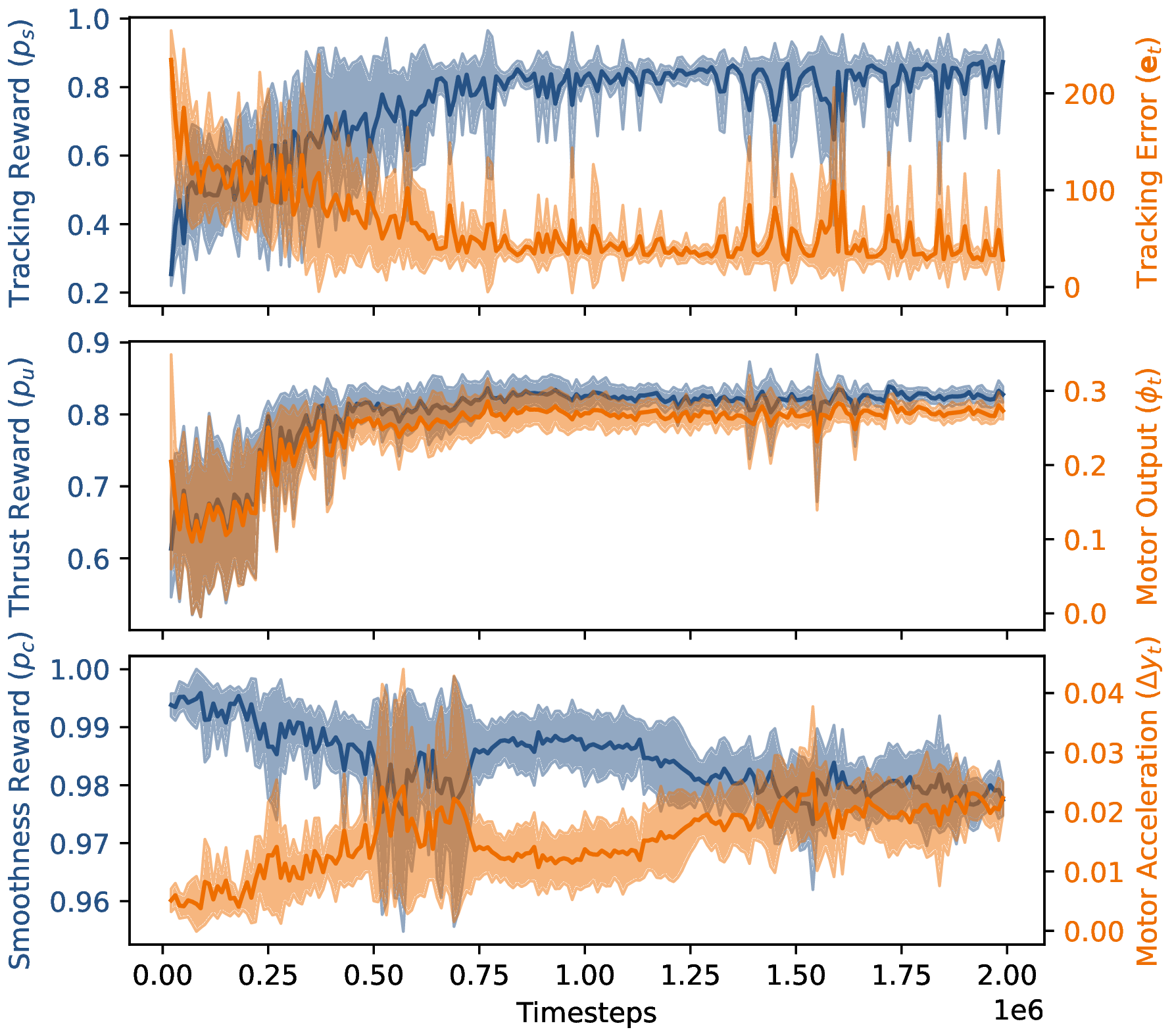}
    \caption{Linear Composition}
    \label{fig:lin_sim}
    \end{subfigure}%
    \begin{subfigure}{.45\textwidth}
        \centering
        \includegraphics[width=\textwidth]{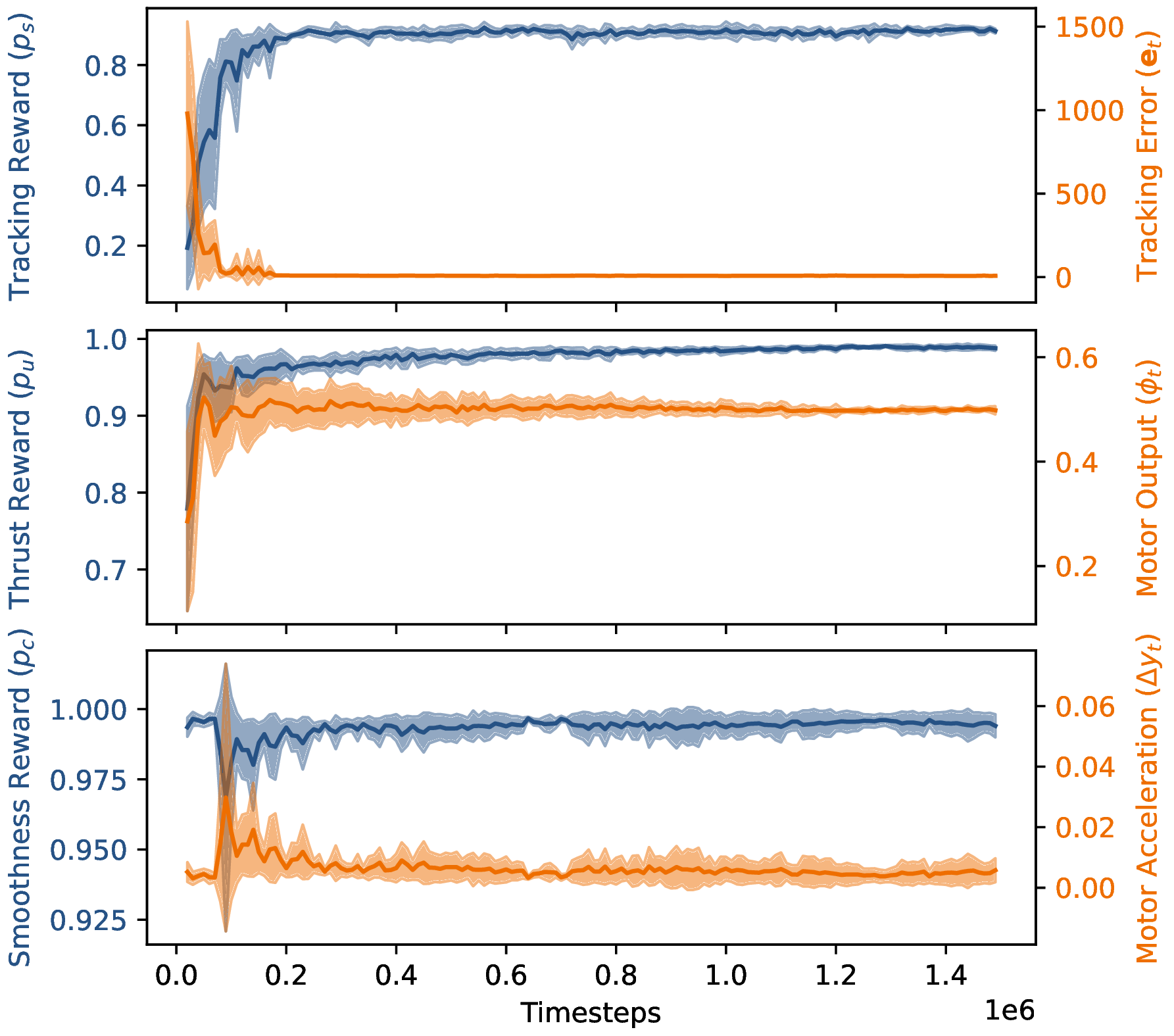}
        \caption{Multiplicative Composition}
        \label{fig:mult_sim}
    \end{subfigure}
    \caption{When comparing multiplicative composition of optimization criteria, as described in Section~\ref{subsec:mult}, against linear composition, we observe that the multiplicative composition quickly and significantly reduces learning variance while achieving comparable or better performance on the tracking error, motor acceleration, and motor actuation objectives.}
    \label{fig:mult_v_lin}
\end{figure*}
    \subsection{Live Adaptation During Unconstrained Flight}\label{subsec:unconstrainedFlight}
    \begin{table}[H]
        \caption{Errors and power consumption in flight before and after 15 steps of live adaptation}
        \vspace{-.5\baselineskip}
        \label{tab:error_amps}
        \centering
        \begin{tabular}{c|c|c}
            \hline & {Before Adaptation} & {After Adaptation} \\
            \hline
            Amperage $\downarrow$ & $13.7 \pm 8.47$ & $7.24 \pm 3.97$\\
            Mean Average Error $\downarrow$ & $12.55 \pm 12.22$ & $14.13 \pm 5.21$\\
            Smoothness $\times 10^{4}$ $\downarrow$ & $12.6 \pm 0.98$ & $5.85 \pm 0.96$ \\
            \hline
        \end{tabular}
        \vspace{-1.5\baselineskip}
    \end{table}
    
    \begin{figure}[H]
        \centering
        \includegraphics[width=\columnwidth]{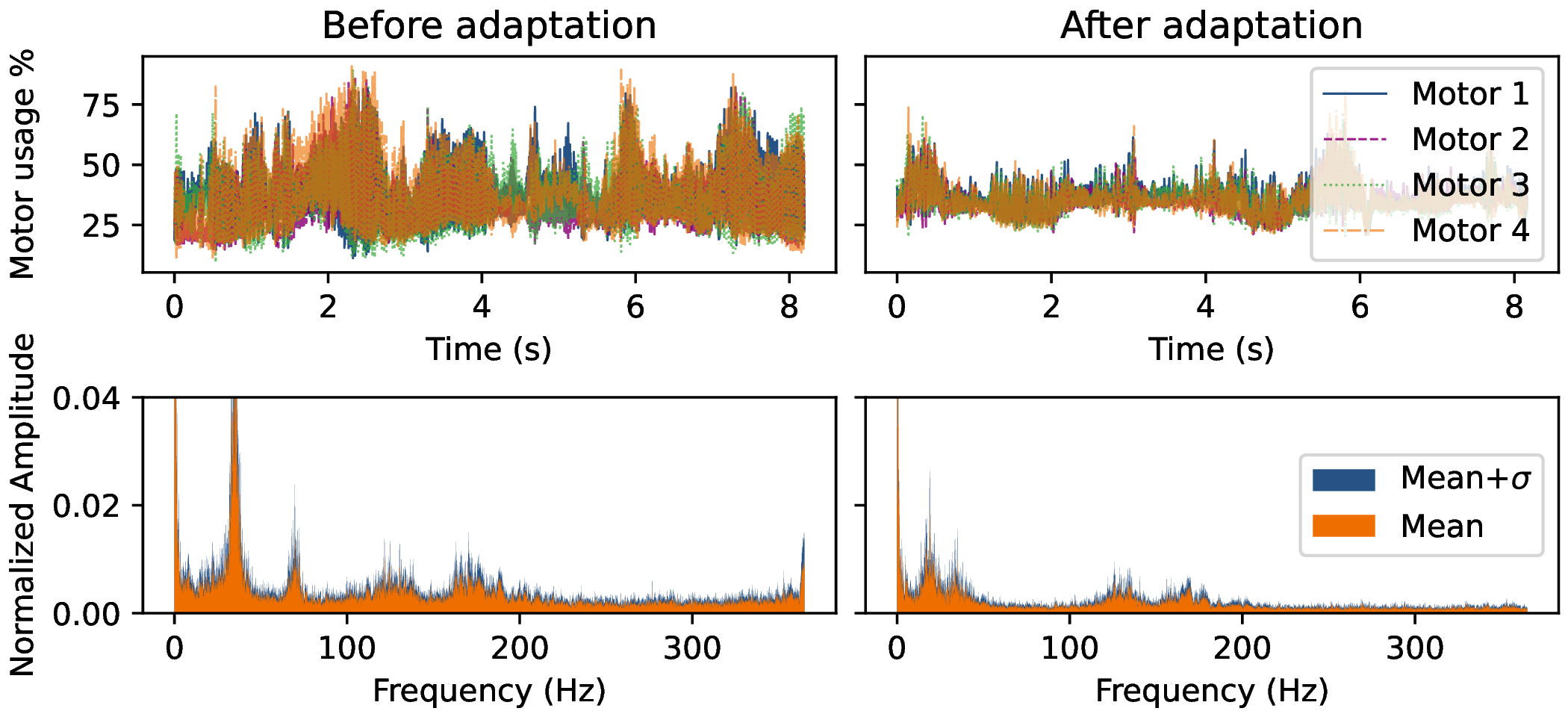}
        \caption{
            Comparisons of motor usage during flight and corresponding FFTs demonstrate the improvement in smoothness after adaptation, as noted from the reduction in high-frequency components on the FFT amplitude plot. 
            While the motor usage plot samples a single trajectory for clarity, the FFTs are averaged over multiple adaptations to demonstrate repeatability.
        }
        \label{fig:fouriers}
        \vspace{-0.5\baselineskip}
    \end{figure}

        We use two key performance metrics in policy evaluation. 
        (1) Tracking accuracy is measured using the Mean Absolute Error (MAE). 
        (2) Smoothness is computed by taking the Fast Fourier Transform (FFT) of the actuation commands, and then computing $Sm$ as defined by Equation 4 in~\cite{mysore2021caps} (lower is better).
        Shown in Fig.~\ref{fig:fouriers} is a comparison of control signals before and after 15 adaptation steps with anchor critics.
        Motor usage is reduced significantly, resulting in improved control smoothness and almost a 50\% reduction in power consumption. Crucially, this is achieved with comparable MAEs to the original simulated agent (Table~\ref{tab:error_amps}), indicating that tracking performance is maintained across the intended operational range, thanks to the anchor preserving the behavior learned on the diverse simulation distribution.
        Table~\ref{tab:error_amps} further details how MAE, smoothness, and current draw change through adaptation.

\section{Applying Multi-Objective Composition Principles from FPL}\label{subsec:mult}

    Linearly combining multiple RL objectives often obscures designer intent. Fulfillment Priority Logic (FPL)~\cite{fpl2025} was introduced to provide principled composition operators for consolidating multi-objective optimization in RL, with one instance being the use of the geometric mean $\pand{0}$ as a smooth AND operator. The FPL $\pand{0}$ operator explicitly targets the simultaneous satisfaction of objectives, represented as objectives with fulfillment levels between $[0,1]$. By making use of this framework, we can scalarize our multi-objective policy optimization as a single objective optimization problem. Since we propose to simulatenously satisfy multiple objectives over both the simulation and real-world domains, we can use the operator provided by FPL to compose our objective functions for both domains into a single scalarized objective function.

    We take the standard expected task fulfillment (normalized Q-value, $Q_{\pi_\theta}$) and the temporal ($L_T$) and spatial ($L_S$) smoothness penalties from CAPS~\cite{mysore2021caps}, along with an actuation penalty term derived from the pre-activation output ($\pi_{\theta}^{:-1}$). These penalty terms are normalized into fulfillment values $F_T, F_S, F_A \in [0,1]$, where higher values represent better fulfillment (or satisfaction of our objectives). The policy optimization then maximizes the FPL conjunction of these fulfillments:
    \begin{equation}
        J_{\pi_\theta} = \pand{0}( Q_{\pi_\theta}(s, a), F_T, F_S, F_A ) \label{eq:mult_opt}. 
    \end{equation}
    This treats all components as fulfillments for the FPL composition. We refer to this DDPG variant as \DDPGx{}.

    As demonstrated empirically (Section~\ref{eval:mult_v_lin}, Fig.~\ref{fig:mult_v_lin}), this FPL-based composition significantly reduces training variance compared to linear approaches.

\subsection{Ablation Study on Applying FPL-based Composition}\label{eval:mult_v_lin}
    To validate the effectiveness of applying FPL-based multiplicative composition shown in Equation~\ref{eq:mult_opt} in our specific DDPG setup, we trained controllers in simulation comparing the linear loss composition from CAPS~\cite{mysore2021caps} against FPL-based multiplicative composition.
    We conducted extensive hyperparameter search with different weighting parameters, both for the CAPS regularization terms as well as the training rewards.
    We show in Fig.~\ref{fig:mult_v_lin} that we achieved higher performance with less tuning and significantly reduced variance using multiplicative composition.
    The results presented in the figure compare 6 independently trained agents for both linear and multiplicative composition.

    \begin{table}
        \centering
        \caption{\rev{Performance comparison of RL algorithms. Training: 300,000 environment invocations, 5 tests per hyperparameter sample. Rewards computed for 10+ samples per algorithm. Step times: Ubuntu v18.04, Intel Core i7-7700 CPU (no GPU optimization).}}
        \rev{
        \begin{tabular}{c|c|c|c|c|c}
            \hline
            Algorithm & DDPG$\times$ & DDPG & PPO (TF2) & SAC & TD3 \\ \hline
            Reward  & $\scriptstyle 0.89\pm0.033$ & $\scriptstyle 0.38\pm0.13$ & $\scriptstyle 0.36\pm0.11$ & $\scriptstyle 0.40\pm0.12$ & $\scriptstyle 0.37\pm0.11$ \\
            Max Rw & $\scriptstyle 0.93$ & $\scriptstyle 0.59$ & $\scriptstyle 0.61$ & $\scriptstyle 0.63$ & $\scriptstyle 0.57$ \\
            Steps (s) & $\scriptstyle 9.6 \times 10^{-3}$ & $\scriptstyle 9.6 \times 10^{-3}$ & $\scriptstyle 3.6 \times 10^{-3}$ & $\scriptstyle 1.44 \times 10^{-2}$ & $\scriptstyle 1.2 \times 10^{-2}$ \\ \hline
        \end{tabular}
        }
        \label{tab:alg_comp}
    \end{table}

    \begin{table}[H]
        \centering
        \caption{\rev{TF2 PPO parameter/reward correlations}}
        \rev{
        \begin{tabular}{c|c|c}
            \hline
            Parameter   & Correlation & Tested Values \\ \hline
            Clip ratio  &  0.098 & [0.0, 0.05, 0.1, 0.2] \\ 
            $\gamma$       & -0.100 & [0.8, 0.9, 0.95, 0.99]\\ 
            Policy LR       &  0.679 & [1e-4, 5e-4, 1e-3, 3e-3, 5e-3] \\ 
            Value LR       &  0.255 & [1e-4, 5e-4, 1e-3, 3e-3, 5e-3] \\
            lam         &  0.056 & [0.9, 0.97, 0.99, 0.995] \\ \hline
        \end{tabular}
        }
        \label{tab:ppo_corr}
    \end{table}
    
    \begin{table}[H]
        \centering
        \caption{\rev{DDPG parameter/reward correlations}}
        \rev{
        \begin{tabular}{c|c|c}
            \hline
            Parameter   & Correlation & Tested Values \\ \hline
            Buffer size  & -0.156 & [1e5, 5e5, 1e6, 5e6] \\
            $\gamma$        & -0.161 & [0.8, 0.9, 0.95, 0.99] \\
            Polyak       &  0.203 & [0.5, 0.9, 0.95, 0.99, 0.995] \\
            Policy LR        & -0.211 & [1e-4, 5e-4, 1e-3, 3e-3, 5e-3] \\
            Critic LR         & -0.676 & [1e-4, 5e-4, 1e-3, 3e-3, 5e-3] \\
            Batch size   & -0.142 & [50, 100, 200, 400] \\
            Action noise    & -0.691 & [0.01, 0.05, 0.1, 0.2] \\ \hline
        \end{tabular}
        }
        \label{tab:ddpg_corr}
    \end{table}
    
    \begin{table}[H]
        \centering
        \caption{\rev{SAC parameter/reward correlations}}
        \rev{
        \begin{tabular}{c|c|c}
            \hline
            Parameter    & Correlation & Tested Values \\ \hline
            Buffer size  & -0.088 & [1e5, 5e5, 1e6, 5e6] \\
            $\gamma$        &  0.272 & [0.8, 0.9, 0.95, 0.99] \\
            Polyak       &  0.102 & [0.5, 0.9, 0.95, 0.99, 0.995] \\
            LR           & -0.256 & [1e-4, 5e-4, 1e-3, 3e-3, 5e-3] \\
            $\alpha$        &  0.356 & [0.01, 0.05, 0.1, 0.2] \\
            Batch size   &  0.866 & [50, 100, 200, 400] \\ \hline
        \end{tabular}
        }
        \label{tab:sac_corr}
    \end{table}
    
    \begin{table}[H]
        \centering
        \caption{\rev{TD3 parameter/reward correlations}}
        \rev{
        \begin{tabular}{c|c|c}
            \hline
            Parameter    & Correlation & Tested Values \\ \hline
            Buffer size  &  0.516 & [1e5, 5e5, 1e6, 5e6] \\
            $\gamma$        &  0.479 & [0.8, 0.9, 0.95, 0.99] \\
            Polyak       &  0.007 & [0.5, 0.9, 0.95, 0.99, 0.995] \\
            Policy LR        & -0.430 & [1e-4, 5e-4, 1e-3, 3e-3, 5e-3] \\
            Critic LR         &  0.023 & [1e-4, 5e-4, 1e-3, 3e-3, 5e-3] \\
            Batch size   &  0.118 & [50, 100, 200, 400] \\
            Act noise    &  0.313 & [0.01, 0.05, 0.1, 0.2] \\ \hline
        \end{tabular}
        }
        \label{tab:td3_corr}
    \end{table}

\section{Hyper-parameter search}\label{A:hypersearch}
    \rev{
    We performed a hyper-parameter search for PPO, DDPG, SAC, and TD3 tailored to the quadrotor attitude control problem.
    The tested parameters and resulting reward correlations are detailed in Tables~\ref{tab:ppo_corr},~\ref{tab:ddpg_corr},~\ref{tab:sac_corr},~and~\ref{tab:td3_corr}, using ranges informed by standard benchmarks.
    }
    \rev{
    Initial comparisons on a simplified tracking task (Table~\ref{tab:alg_comp}) indicated a performance advantage for our proposed \DDPGx{} variant over standard implementations after 300,000 training steps.
    Observations of standard algorithms revealed challenges specific to this control domain: TF2 implementations of PPO and TD3 often resulted in saturated motor commands, hindering effective attitude control, while SAC exhibited slower learning compared to DDPG.
    Vanilla DDPG showed potential for nuanced control but suffered from high training variance.
    }
    \rev{
    These findings motivated the use of \DDPGx{}, which integrates DDPG with the FPL-based multiplicative composition described in Section~\ref{subsec:mult}.
    This approach effectively mitigated the observed variance issues, yielding consistent, high-performance policies suitable for real-world flight.
    Agents achieving average rewards above 0.8 in simulation reliably demonstrated stable flight characteristics in our hardware tests.
    }

\section{Known Limitations}\label{subsec:limitations}

(1) In cases where domain gaps are large, the very stability offered by the anchor critics may significantly compromise fine-tuning of behavior and could result in policies that are ill-suited for both the initial and transferred domain.

(2) While anchors clearly reduce power consumption without resulting in catastrophic forgetting, the adapted agents were not observed to improve their tracking performance significantly, even though this was one of the key learning objectives of the policy optimization. We hypothesize that this is due to the fact that we relied on replay buffers from simulation to evaluate anchor critics, rather than have a continuously updated anchor via a simulation in the loop, we deem this a promising direction for future work.

(3) The complexity of implemention may detract from the wide-applicability of the approach, even though we provide a generic implementation applicable over a range of robotic platforms via \firmware{} and \framework{}, orchestrating data from simulation and real-world data is non-trivial.

\section{Acknowledgements}
    We thank Patrick Carpanedo for his contributions to the video presentation and adding polishing touches to the manuscript.

\bibliographystyle{IEEEtran}
\bibliography{references.bib}


\end{document}